\documentclass[sigconf]{acmart}

\renewcommand\footnotetextcopyrightpermission[1]{} 
\hypersetup{draft}

\usepackage{url}            
\usepackage{amsfonts}       
\usepackage{nicefrac}       
\usepackage{microtype}      
\usepackage{xcolor}

\usepackage{bbm}
\usepackage{multirow}
\usepackage{hhline}
\usepackage{graphicx}
\graphicspath{ {figures/} }
\usepackage{subfig}
\usepackage{wrapfig}
\usepackage{amsmath}
\usepackage{breqn}
\usepackage[ruled]{algorithm2e}
\usepackage[utf8]{inputenc}
\usepackage{array}
\newcolumntype{C}[1]{>{\centering\arraybackslash}m{#1}}

\def\BibTeX{{\rm B\kern-.05em{\sc i\kern-.025em b}\kern-.08emT\kern-.1667em\lower.7ex\hbox{E}\kern-.125emX}}
    
\acmConference[DAC '19]{The 56th Annual Design Automation Conference 2019}{June 2--6, 2019}{Las Vegas, NV, USA}

\begin{document}

%
\title{FLightNNs: Lightweight Quantized Deep Neural Networks for Fast and Accurate Inference}
%
\author{Ruizhou Ding, Zeye Liu, Ting-Wu Chin, Diana Marculescu, and R. D. (Shawn) Blanton}
\email{{rding,zeyel,tingwuc,dianam,rblanton}@andrew.cmu.edu}
\affiliation{%
  \institution{Carnegie Mellon University, Pittsburgh, U.S.A.}
}

%
\renewcommand{\shortauthors}{Ding, et al.}

%
\begin{abstract}
To improve the throughput and energy efficiency of Deep Neural Networks (DNNs) on customized hardware, lightweight neural networks constrain the weights of DNNs to be a limited combination (denoted as $k\in\{1,2\}$) of powers of 2. In such networks, the multiply-accumulate operation can be replaced with a single shift operation, or two shifts and an add operation. 
To provide even more design flexibility, the $k$ for each convolutional filter can be optimally chosen instead of being fixed for every filter. In this paper, we formulate the selection of $k$ to be differentiable, and describe model training for determining $k$-based weights on a per-filter basis. Over 46 FPGA-design experiments involving eight configurations and four data sets reveal that lightweight neural networks with a flexible $k$ value (dubbed FLightNNs) fully utilize the hardware resources on Field Programmable Gate Arrays (FPGAs), our experimental results show that FLightNNs can achieve 2$\times$ speedup when compared to lightweight NNs with $k=2$, with only 0.1\% accuracy degradation. Compared to a 4-bit fixed-point quantization, FLightNNs achieve higher accuracy and up to 2$\times$ inference speedup, due to their lightweight shift operations. In addition, our experiments also demonstrate that FLightNNs can achieve higher computational energy efficiency for ASIC implementation. 

\end{abstract}

%
%


%
\maketitle

\makeatletter
\patchcmd{\@maketitle}
  {\addvspace{0.5\baselineskip}\egroup}
  {\addvspace{-1\baselineskip}\egroup}
  {}
  {}
\makeatother

\fancyfoot{}
\fancyfoot[C]{\thepage}
\section{Introduction}
Emerging vision, speech and natural language applications have widely adopted deep learning models and, as a result, have achieved state-of-the-art accuracy. Furthermore, recent industrial efforts have focused on implementing the models on mobile devices~\cite{NNAPI}. However, real-time applications based on these deep models may incur unacceptably large latencies and can easily drain the battery on energy-limited devices. For example, smartphones can only run the AlexNet-based object detection for one hour~\cite{yang2017designing}. Therefore, prior research has proposed model compression techniques including pruning and quantization to satisfy the stringent energy and speed requirements~\cite{lin2016fixed}. 

\begin{figure}[t]
\centering
    \includegraphics[width=0.35\textwidth]{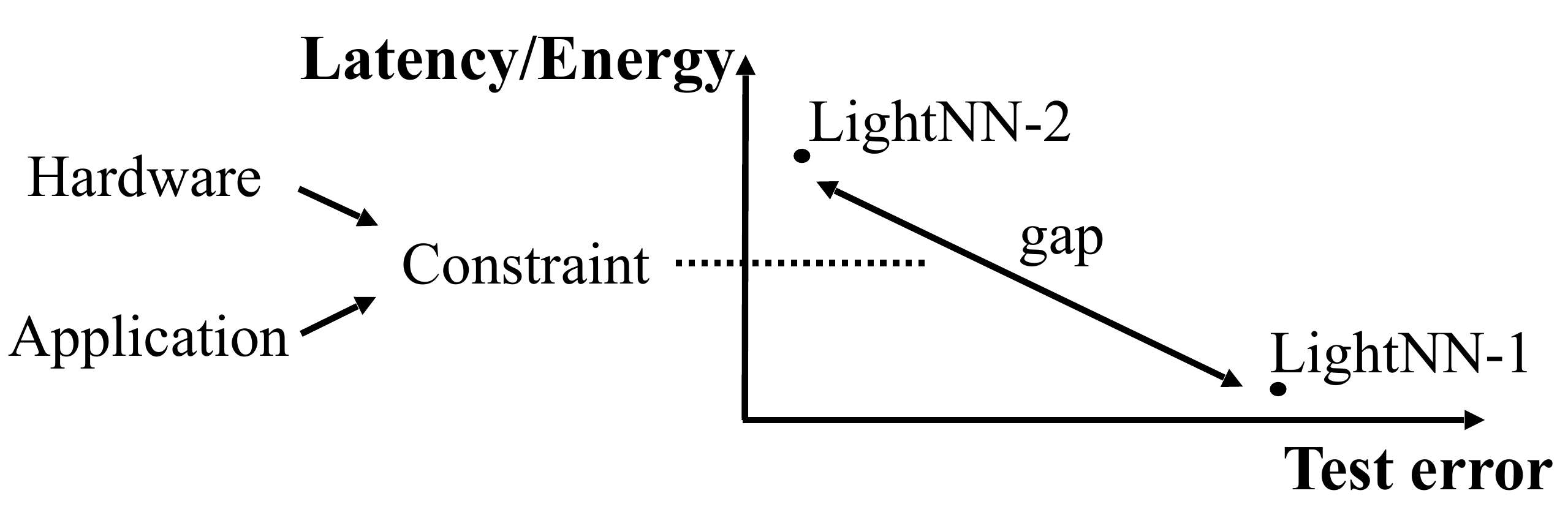}
    \vspace{-12pt}
    \caption{A discrete Pareto-optimal curve for LightNN models $w.r.t.$ test error and latency/energy. More continuous Pareto-optimal points are needed to adapt to the latency/energy constraints determined by the hardware and application.}\label{fig:gaps}
\vspace{-18pt}
\end{figure}

One of the recently proposed quantization approaches, LightNN, constrains the weights of DNNs to be a sum of $k$ powers of 2, and therefore can use shift and add operations to replace the multiplications between activations and weights~\cite{ding2017lightnn}. For LightNN-1\footnote{LightNN-$k$ quantizes weights to be the sum of $k$ powers of 2.}, all the multiplications of the DNNs will be replaced by a shift operation, while for LightNN-2, two shifts and an add replace the multiplication. Since shift operations are much more lightweight on customized hardware (\textit{e.g.}, FPGA or ASIC), LightNNs can achieve faster speed and lower energy consumption, and generally maintain accuracy for over-parameterized models~\cite{ding2017lightnn,ding2018lightening}. Although LightNNs provide better energy-efficiency, they lack the flexibility to provide fine-grained trade-offs between energy and accuracy. As shown in Fig.~\ref{fig:gaps}, the energy efficiency for these models also exhibits gaps, making the Pareto front of accuracy and energy discrete. However, a continuous accuracy and energy/latency trade-off is an important feature for designers to target different market segments (\emph{e.g.}, IoT devices, edge devices, and mobile devices).

To provide a more flexible Pareto front for the LightNN framework, we propose to equip each convolutional filter with the freedom to use a different number of shift-and-add operations to approximate multiplications. Specifically, we introduce a set of free variables $\mathbf{k}=\{\mathbf{k}_1,\dots,\mathbf{k}_F\}$ where each element represents the number of shift-and-add for the corresponding convolutional filter. As a result, a more contiguous Pareto front can be achieved. For example, if we constrain $\mathbf{k}\in\{1,2\}^F$, then the throughput and energy consumption of the new model will sit between LightNN-1 ($\mathbf{k}=\{1\}^F$) and LightNN-2 ($\mathbf{k}=\{2\}^F$). Formally, we are solving $\min_{\mathbf{w},\mathbf{k}}\mathcal{L}(\mathbf{w},\mathbf{k})$, where $\mathcal{L}$ is the loss function and $\mathbf{w}$ is the weights vector. However, the commonly adopted stochastic gradient descent (SGD) algorithm does not apply in this case since $\mathcal{L}$ is non-differentiable $w.r.t.$ $\mathbf{k}$. In this paper, we propose a \textit{differentiable training algorithm} which enables end-to-end optimization with standard SGD. The resulting network is dubbed \emph{\textbf{F}LightNN} for its flexible $\mathbf{k}$ values.

\section{Related Work}
Prior work has extensively explored approaches to reduce latency and energy consumption of DNNs on hardware, through both algorithmic~\cite{hubara2016binarized,yang2017designing} and hardware~\cite{chen2017eyeriss,zhao2018processing} efforts. Since the latency and energy consumption of DNNs generally stem from computational cost and memory accesses, prior work in the algorithmic domain mainly focuses on the reduction of FLOPs and model size. Some work reduces the number of parameters through weight pruning~\cite{han2015learning}, while some other work introduces structural sparsity via filter pruning for Convolutional Neural Networks (CNNs)~\cite{wen2016learning} to enable speedup on general hardware platforms incorporating CPUs and GPUs. To reduce the model size, previous work has also conducted neural architecture search with energy constraint~\cite{yang2017designing,stamoulis2018hyperpower,marculescu2018hardware,stamoulis2018designing}. In addition to algorithmic advances, prior art has also proposed methodologies to achieve fast and energy-efficient DNNs. Some previous work proposes the co-design of the hardware platform and the architecture of the neural network running on it~\cite{brooks2018co}. Some work proposes more lightweight DNN units for faster inference on general-purpose hardware~\cite{sandler2018mobilenetv2}, while others propose hardware-friendly DNN computation units to enable energy-efficient implementation on customized hardware~\cite{tann2017hardware}.

By reducing the weight and activation precision, DNN quantization has proved to be an effective technique to improve the speed and energy efficiency of DNNs on customized hardware, due to its lower computational cost and fewer memory accesses~\cite{gupta2015deep}. Gupta \textit{et al.} show that a DNN with 16-bit fixed-point representation can achieve competitive accuracy compared to the full-precision network~\cite{gupta2015deep}. In the same vein, Zhou \textit{et al.} explored the DNN accuracy \textit{w.r.t.} a wide range of bit widths~\cite{zhou2016dorefa}. These uniform quantization approaches enable fixed-point hardware implementation for DNNs. Courbariaux \textit{et al.} propose BinaryConnect, which uses only 1 bit for the DNN parameters, turning multiplications into XNOR operations on customized hardware~\cite{courbariaux2015binaryconnect}. However, these models require an over-parameterized model size to maintain a high accuracy~\cite{ding2017lightnn}. 

LightNNs constrain the model weights to be a power of 2, or the sum of a limited number of powers of 2~\cite{ding2017lightnn}, while the activations use fixed-point quantization. Therefore, the multiplication between weights and activations can be implemented in hardware by shift operations and fixed-point additions. Compared to DNNs with fixed-point quantization, LightNNs replace the fixed-point multipliers by more lightweight shift operators, or shift and additions. Since the shift operators can be implemented using Look-Up Table (LUT) on FPGA while fixed-point multipliers require Digital Signal Processing (DSP) units, LightNNs can have higher inference speed than fixed-point DNNs when run on DSP-bounded FPGAs. In addition, in an ASIC implementation, shift operations are more lightweight than multiplications, making LightNNs more energy and area efficient than fixed-point DNNs. 

However, LightNNs use a single $k$ value (\textit{i.e.}, the number of shifts per multiplication) across the whole network, and therefore lack flexibility to provide a fine-grained energy/latency and accuracy trade-off for hardware designers. Therefore, we propose FLightNNs which use customized $k$ values for each convolutional filter to enable a more continuous Pareto front. Recent work has explored the idea of differentiable training for architecture search~\cite{liu2018darts} and neural network pruning~\cite{louizos2018learning}. In this paper, we propose an end-to-end differentiable training algorithm for FLightNNs via approximate gradient computation for non-differentiable operations and regularization to encourage sparsity. Moreover, the proposed differentiable training approach uses gradual quantization, which can achieve higher accuracy than LightNN-1 without increasing latency. In summary, this paper has the following key contributions:

(i) We propose a differentiable training algorithm for FLightNNs, which provides a continuous Pareto front for hardware designers to search for a highly accurate model under the hardware resource constraints. 

(ii) The differentiable training for FLightNNs enables gradual quantization, and further pushes forward the Pareto-optimal curve.





\section{LightNN Overview}
As a quantized DNN model, LightNNs constrain the weights of a network to be the sum of $k$ powers of 2, denoted as LightNN-$k$. Thus, the multiplications between weights and activations can be implemented with $k$ shift operations and $k-1$ additions. Specifically, LightNN-1 constrains the weights to be a power of 2, and only uses a shift for a multiplication. The approximation function used by LightNN-$k$ to quantize a full-precision weight $w$ can be formulated in a recursive way: $\mathcal{Q}_k(w)=\mathcal{Q}_{k-1}(w)+\mathcal{Q}_{1}(w-\mathcal{Q}_{k-1}(w))$ for $k>1$, where $\mathcal{Q}_1(w)=sign(w)\times 2^{[log(|w|)]}$ which rounds the weight $w$ to a nearest power of 2.

LightNNs are trained with a modified backpropagation algorithm. In the forward phase of each training iteration, the parameters are first approximated using the $\mathcal{Q}_k$ function. Then, in the backward phase, the gradients of loss \textit{w.r.t.} quantized weights are computed, and applied to the full-precision weights in the weight update phase. LightNNs have been proved to be accurate and energy-efficient on customized hardware~\cite{ding2017lightnn}. LightNN-2 can generally have an accuracy close to full-precision DNNs, while LightNN-1 can achieve higher energy efficiency than LightNN-2. Due to the nature of the discrete $k$ values, there exists a gap between LightNN-1 and LightNN-2 \textit{w.r.t.} accuracy and energy. We propose to customize the $k$ values for each convolutional filter, and thus, achieve a smoother energy-accuracy trade-off to provide hardware designers with more design options.

\section{Differentiable Training for FLightNNs}

In this section, we first define the quantization function, and then introduce the end-to-end training algorithm for FLightNNs, equipped with a regularization loss to penalize large $\mathbf{k}$ values.

\subsection{Quantization function}
\begin{figure}[t]
  \centering
  \includegraphics[width=0.47\textwidth]{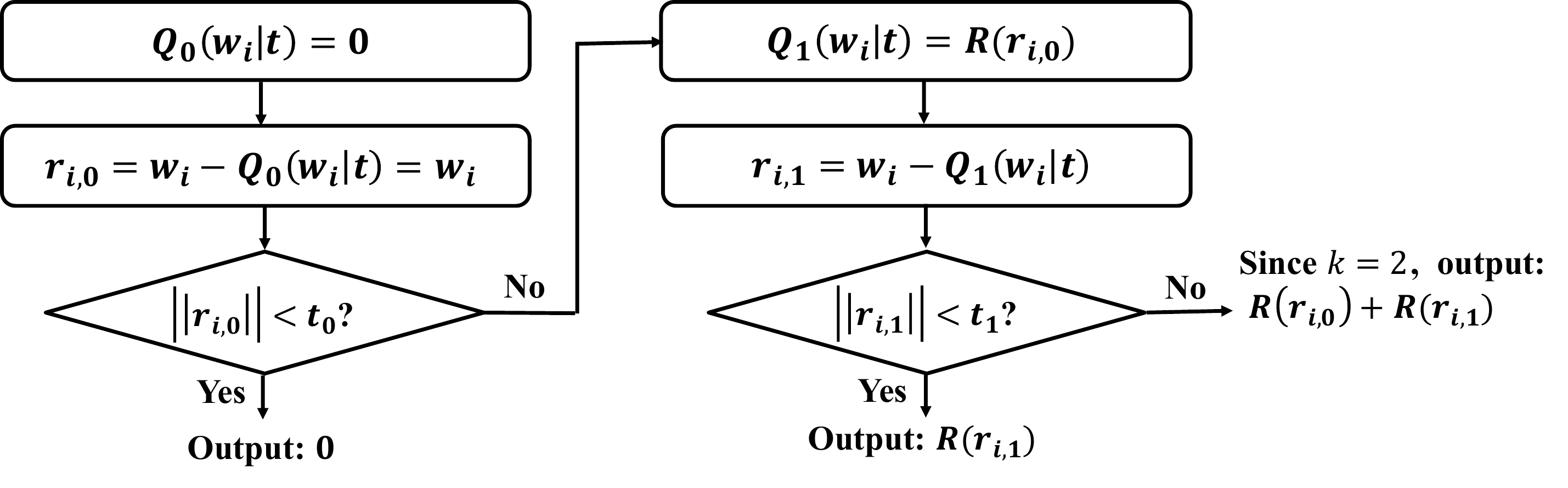}
  \vspace{-12pt}
  \caption{Quantization flow for $k=2$.}\label{fig:quantization-flow}
  \vspace{-12pt}
\end{figure}

We first denote the $i^{th}$ filter of the network as $\mathbf{w}_i$ and the quantization function for the filter $\mathbf{w}_i$ as $\mathcal{Q}_k(\mathbf{w}_i|\mathbf{t})$, where $k=\max_i\mathbf{k}$ is the maximum number of shifts used for this network, and vector $\mathbf{t}$ is a latent variable that controls the approximation (\textit{e.g.}, some threshold value). Also, we denote the residual resulting from the approximation as $\mathbf{r}_{i,k}=\mathbf{w}_i-\mathcal{Q}_k(\mathbf{w}_i|\mathbf{t})$. Then, we formally define the quantization function as follows:
\vspace{-3pt}
  \[
    \mathcal{Q}_k(\mathbf{w}_i|\mathbf{t}) =\left\{
                \begin{array}{ll}
                  0,~~~~~~~~~~~~~~~~~~~~~~~~~~~~~~~~~~~~~~~~~~~~~~~\textup{if}~k=0\\
                  \sum_{j=0}^{k-1} \mathbbm{1}(||\mathbf{r}_{i,j}||_2>\mathbf{t}_j)R(\mathbf{r}_{i,j}),~\textup{if}~k\ge1
                \end{array}
              \right.
\vspace{-3pt}
  \]
where $R(x)=sign(x)\times2^{[log(|x|)]}$ rounds the input variable to a nearest power of 2, and $[.]$ is a rounding-to-integer function. This quantization flow is shown in Fig.~\ref{fig:quantization-flow}. To interpret the thresholds $\mathbf{t}$, $\mathbf{t}_0$ determines whether this filter is pruned out, and $\mathbf{t}_1$ determines whether one shift is enough, \textit{etc}. Then, the number of shifts for the $i$-th filter is $\mathbf{k}_i=\sum_{j=0}^{k-1}\mathbbm{1}(||\mathbf{r}_{i,j}||_2>\mathbf{t}_j)$. Therefore, choosing $\mathbf{k}_i$ per filter is equivalent to finding optimal thresholds $\mathbf{t}$.

\begin{figure}[t]
  \centering
  \includegraphics[width=0.47\textwidth]{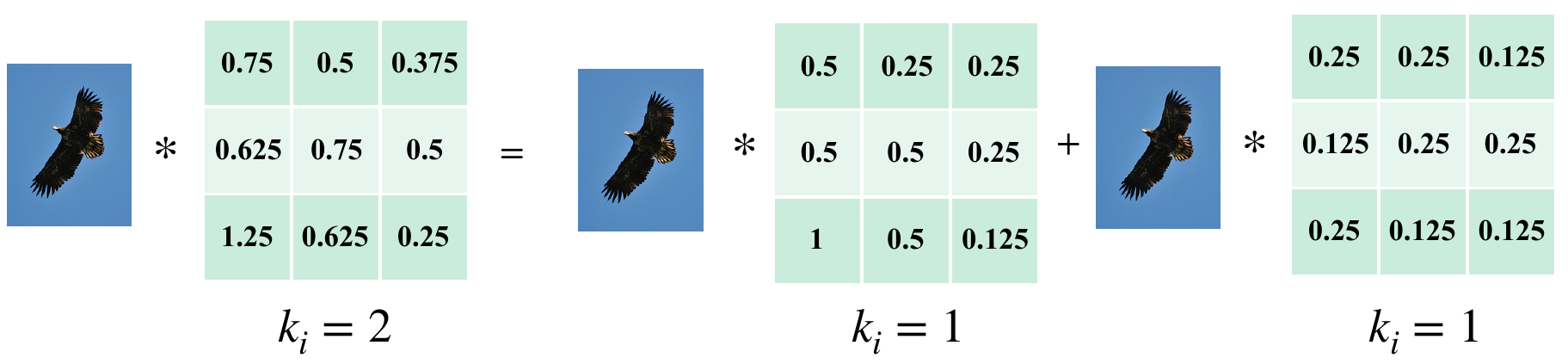}
  \vspace{-12pt}
  \caption{Equivalent conversion from a convolution with a $\mathbf{k}_i>1$ filter to $\mathbf{k}_i$ convolutions each with a $\mathbf{k}_i=1$ filter. This transforms the hardware implementation of the FLightNN into LightNN-1.}\label{fig:decompose-k}
  \vspace{-9pt}
\end{figure}

The FLightNN quantization approach targets efficient hardware implementation. Instead of assigning a customized $\mathbf{k}_i$ for each weight, FLightNNs have customized $\mathbf{k}_i$ values per filter, and therefore preserve the structural sparsity. As shown in Fig.~\ref{fig:decompose-k}, the convolution with a $\mathbf{k}_i=2$ filter can be equivalently converted to the sum of two convolutions each with a $\mathbf{k}_i=1$ filter. Thus, FLightNNs can be efficiently implemented as LightNN-1 with an extra summation of feature maps per layer. 

\subsection{Differentiable training}
Instead of picking the threshods $t$ by hand, we consider them as trainable parameters. Therefore, the loss function $\mathcal{L}(\mathbf{w}\footnote{The bias term is omitted for simplicity.},\mathbf{t})$ is a function of both weights and thresholds. Similar to prior work on DNN quantization~\cite{zhou2016dorefa,courbariaux2015binaryconnect}, we use the straight-through estimator (STE)~\cite{bengio2013estimating} to compute $\frac{\partial\mathcal{L}}{\partial \mathbf{w}_i}$. By defining $\frac{\partial \mathbf{w}_i^q}{\partial \mathbf{w}_i}=1$ where $\mathbf{w}_i^q=\mathcal{Q}_k(\mathbf{w}_i|\mathbf{t})$ is the quantized $\mathbf{w}_i$; therefore, we have $\frac{\partial\mathcal{L}}{\partial \mathbf{w}_i}=\frac{\partial\mathcal{L}}{\partial \mathbf{w}_i^q}\cdot\frac{\partial \mathbf{w}_i^q}{\partial \mathbf{w}_i}=\frac{\partial\mathcal{L}}{\partial \mathbf{w}_i^q}$, which becomes a differentiable expression. 

To compute the gradient for thresholds, \textit{i.e.}, $\frac{\partial \mathbf{w}_i^q}{\partial \mathbf{t}_j}$, we relax the indicator function $g(x,\mathbf{t}_j)=\mathbbm{1}(x>\mathbf{t}_j)$ to a sigmoid function~\cite{han1995influence}, $\sigma(.)$, when computing gradients, \textit{i.e.}, $\hat{g}(x,\mathbf{t}_j)=\sigma(x-\mathbf{t}_j)$. In addition, we use STE to compute the gradient for $R(x)$. Thus, the gradient $\frac{\partial \mathbf{w}_i^q}{\partial \mathbf{t}_j}$ can be computed by:
\begin{equation*}
\begin{split}
    & \frac{\partial \mathcal{Q}_{\mathbf{k}_i}(\mathbf{w}_i|\mathbf{t})}{\partial \mathbf{t}_j} = \sum_{l=0}^{\mathbf{k}_i-1}\frac{\partial\sigma(||\mathbf{r}_{i,l}||_2-\mathbf{t}_l)}{\partial \mathbf{t}_j}R(\mathbf{r}_{i,l}) + \sigma(||\mathbf{r}_{i,l}||_2-\mathbf{t}_l)\frac{\partial R(\mathbf{r}_{i,l})}{\partial \mathbf{t}_j} \\
    = & \sum_{l=0}^{\mathbf{k}_i-1}\sigma'(||\mathbf{r}_{i,l}||_2-\mathbf{t}_l)
    (\frac{\partial ||\mathbf{r}_{i,l}||_2}{\partial \mathbf{t}_j} - \frac{\partial \mathbf{t}_l}{\partial \mathbf{t}_j})
    R(\mathbf{r}_{i,l}) +
    \sigma(||\mathbf{r}_{i,l}||_2-\mathbf{t}_l)\frac{\partial \mathbf{r}_{i,l}}{\partial \mathbf{t}_j} 
\end{split}
\end{equation*}
where $\frac{\partial ||\mathbf{r}_{i,l}||_2}{\partial \mathbf{t}_j}$ and $\frac{\partial \mathbf{r}_{i,l}}{\partial \mathbf{t}_j}$ are $0$ for $l<j$; otherwise, they can be computed with the result of $\frac{\partial \mathcal{Q}_{l}(\mathbf{w}_i|\mathbf{t})}{\partial \mathbf{t}_j}$. $\frac{\partial \mathbf{t}_l}{\partial \mathbf{t}_j}=\mathbbm{1}(l=j)$.

\subsection{Regularization}
\begin{figure}[t]
  \centering
  \includegraphics[width=0.43\textwidth]{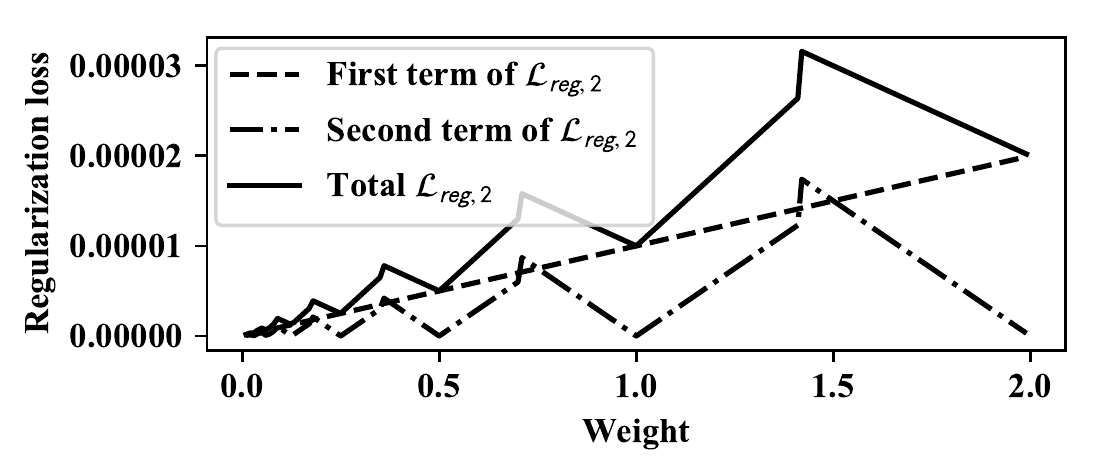}
  \vspace{-12pt}
  \caption{Regularization loss curve $w.r.t.$ weight value.}\label{fig:loss}
  \vspace{-9pt}
\end{figure}

To encourage smaller $\mathbf{k}_i$ for the filters, we also add a regularization loss:
    $\mathcal{L}_{reg,k}(\mathbf{w}) = \sum_{j=0}^{k-1}\lambda_j\sum_i||\mathbf{r}_{i,j}||_2$
where $\lambda_j$ performs as a handle to balance accuracy and model sparsity. This regularization loss is the sum of several group Lasso losses, since they can introduce structural sparsity~\cite{wen2016learning}. The first item $\lambda_0\sum_i||\mathbf{r}_{i,0}||_2=\lambda_0\sum_i||\mathbf{w}_i||_2$ is used to prune the whole filters out, while the other items ($j>0$) regularize the residuals. Fig.~\ref{fig:loss} shows the two items of regularization loss and their sum for the case $k=2$, with $\lambda_0$=1e-5 and $\lambda_1$=3e-5. Therefore, the total loss for training a FLightNN is: $\mathcal{L}_{total}(\mathbf{w}, \mathbf{t})=\mathcal{L}_{CE}(\mathbf{w}, \mathbf{t}) + \mathcal{L}_{reg,k}(\mathbf{w})$.

The new training algorithm is summarized in Algo.~\ref{algo:FLightNN-training}. This is the same as the conventional backpropagation algorithm for full-precision DNNs, except that in the forward phase, the weights are quantized given the thresholds $\mathbf{t}$. Then, due to the differentiability of the quantization function \textit{w.r.t.} $\mathbf{w}$ and $\mathbf{t}$, one can compute their gradients and update their values in each training iteration. 

\begin{algorithm}[t]
 \textbf{Input}: Training dataset (${\textbf{x}}$, ${\textbf{y}}$), where ${\textbf{x}}$ is input and ${\textbf{y}}$ is label; parameters after the $(p-1)$-th iteration: ${\textbf{w}}_{p-1}$ (weights), ${\textbf{b}}_{p-1}$ (biases), and quantization thresholds ${\textbf{t}}_{p-1}$; quantization function $\mathcal{Q}_k({\textbf{w}}|{\textbf{t}})$; DNN forward computation function $g(x,w,b)$; maximum $\textit{k}$ value used for all filters; regularization loss coefficients $\lambda$; learning rate $\eta$.
 
 \textbf{Output}: Updated weights ${\textbf{w}}_{p}$, biases ${\textbf{b}}_{p}$ and thresholds ${\textbf{b}}_{p}$.

 \For{each mini-batch of ${\textbf{x,~y}}$} {
  1. \textbf{Quantize weights}: ${\textbf{w}}^q=\mathcal{Q}_k ({\textbf{w}}_{p-1}|{\textbf{t}}_{p-1})$ \\
  2. \textbf{Forward}: compute intermediate results and cross entropy loss function $\mathcal{L}_{CE}$ with $g(\cdot)$, ${\textbf{w}}^q$, ${\textbf{b}}_{p-1}$, and mini-batch of ${\textbf{x}}$; compute regularization loss $\mathcal{L}_{reg,k}$ with $\lambda$ and ${\textbf{w}}_{p-1}$; get the total loss $\mathcal{L}_{total} = \mathcal{L}_{CE} + \mathcal{L}_{reg,k}$ \\
  3. \textbf{Backward}: compute derivatives $\frac{\partial \mathcal{L}_{total}}{\partial {\textbf{w}}^q}$, $\frac{\partial \mathcal{L}_{total}}{\partial {\textbf{b}}_{p-1}}$, and $\frac{\partial \mathcal{L}_{total}}{\partial {\textbf{t}}_{p-1}}$ \\
  4. \textbf{Update parameters}: ${\textbf{w}}_{p} = {\textbf{w}}_{p-1}-\eta\frac{\partial \mathcal{L}_{total}}{\partial {\textbf{w}}^q}$; ${\textbf{b}}_{p} = {\textbf{b}}_{p-1}-\eta\frac{\partial \mathcal{L}_{total}}{\partial {\textbf{b}}_{p-1}}$; ${\textbf{t}}_{p} = {\textbf{t}}_{p-1}-\eta\frac{\partial \mathcal{L}_{total}}{\partial {\textbf{t}}_{p-1}}$ \\
 } 
 \caption {FLightNN Training Epoch}\label{algo:FLightNN-training}
\end{algorithm}

\section{Experimental Results}
In this section, we first introduce the experiment setup. Then, we show the accuracy results of different quantized DNN models by software training, as well as their throughput on the FPGA and energy efficiency on the ASIC, to verify the effectiveness of FLightNNs. 

\subsection{Setup}
We conduct experiments on both small and large CNNs for CIFAR-10, SVHN, CIFAR-100 and ImageNet datasets. The eight adopted network configurations are shown in Table~\ref{table:network-settings}. To explore the FLightNN performance on different types of network structures, we use a VGG structure with a series of stacked convolutional layers for Network 1, 3, 4 and 5, and adopt the ResNet structure with skip connections across layers for network 2, 6, 7 and 8. Networks 1, 2 and 3 are used for experiments on CIFAR-10; networks 4 and 5 are used for SVHN; networks 6 and 7 are used for CIFAR-100; the last one, network 8, is used for ImageNet. 
For all networks, each convolutional layer is followed by a batch normalization layer and a Leaky ReLU activation function~\cite{maas2013rectifier}, and optionally followed by a max-pooling layer. We use the Adam optimizer~\cite{kingma2014adam} to train the network. For each of the networks, we train different quantized models including full-precision DNNs, fixed-point DNNs with 4-bit weights and 8-bit activations, LightNN-2 with 8-bit weights and 8-bit activations, LightNN-1 with 4-bit weights and 8-bit activations, and FLightNNs with 8-bit activations. Due to large training times and limitations in computing resources, we train the ImageNet dataset on a ResNet-10 with reduced width (\textit{i.e.}, network 8), for LightNN-1, LightNN-2 and FLightNNs. For all FLightNNs, we initialize the threshods $\mathbf{t}$ to $\mathbf{0}$, and set the largest shifts $k$ as 2. For all, except the 32-bit full-precision model, we use 8-bit fixed-point quantization for the activations. By varying $\lambda$, we can have different accuracy-throughput or accuracy-energy trade-offs for FLightNNs. All these networks are trained in software through PyTorch.


\begin{table}[t]
\centering
\caption{Network settings. ``Depth" is the number of convolutional layers in the network. ``Width" is the number of convolutional filters of the largest layer.}
\vspace{-12pt}
\label{table:network-settings}
\begin{tabular}{C{1.5cm}|C{1.5cm}|C{1.5cm}|C{0.9cm}|C{0.9cm}}
    \hline
    Network ID & Parameters & Structure & Depth & Width  \\
    \hline
    1 & 0.08M & VGG & 7 & 64 \\
    \hline
    2 & 0.7M & ResNet & 18 & 128 \\
    \hline
    3 & 4.6M & VGG & 7 & 512 \\
    \hline
    4 & 0.03M & VGG & 4 & 64 \\
    \hline
    5 & 0.1M & VGG & 4 & 128 \\
    \hline
    6 & 0.7M & ResNet & 18 & 128 \\
    \hline
    7 & 2.8M & ResNet & 18 & 256 \\
    \hline
    8 & 1.8M & ResNet & 10 & 256 \\
    \hline
\end{tabular}
\end{table}

\subsection{Accuracy-throughput trade-off on FPGA}\label{sec:fpga}
To show the accuracy-throughput trade-off of the models, we implement the inference of each network's largest convolutional layer for each of the quantized DNN models on FPGA since prior work has shown that convolution operations typically take over 90\% of the computation time of a CNN~\cite{Jason_FPGA}. Our implementation is built on the Xilinx Zynq ZC706 evaluation board. Its working frequency is 100 MHz. Pre-synthesis is executed on an Intel i7-4790 CPU (3.6GHz) with 16GB RAM. We use Vivado HLS~\cite{vivado1} for FPGA implementation. The C code of DNN designs are parallelized by adding HLS-defined pragma and the parallel version is validated with the Vivado HLS timing analysis tool. To make a fair comparison, the same pragma and directives are used for full-precision, fixed-point DNNs, LightNNs and FLightNNs, and we follow the same scheduling settings as prior work~\cite{ding2018lightening}. Batched inference is adopted, and the maximum batch size without running out of FPGA resources is set to obtain the highest throughtput. 

Tables~\ref{table:results_cifar10}, \ref{table:results_svhn}, \ref{table:results_cifar100} and \ref{table:results_imagenet} show the accuracy and throughput comparison for full-precision DNNs, fixedpoint DNNs, LightNNs and FLightNNs. For all the experimented datasets, LightNNs show the advantage of flexible accuracy-speed trade-offs. In most of the networks (\textit{e.g.}, networks 1, 3, 6 and 7), FLightNNs can achieve an accuracy close to LightNN-2, but have much higher speedup than LightNN-2. Thus, FLightNNs provide continuous trade-offs for accuracy and speed. Compared to the fixed-point quantization, FLightNNs can achieve higher accuracy, and up to $2.0\times$, $1.8\times$ and $1.8\times$ speedup for CIFAR-10, SVHN, CIFAR-100 datasets, respectively. This is because the multiplication is replaced by shift operators, which require only LUT resources on FPGA while the multipliers require DSP units which are generally more scarce than LUT. Therefore, the computation for FLightNNs allows larger batch sizes than that of fixed-point DNNs, increasing data parallelism, and thus, improving the throughput. 

It is also interesting to note that by comparing some FLightNNs (\textit{e.g.}, FL$_{1a}$, FL$_{2a}$, FL$_{3a}$, FL$_{6a}$ and FL$_{7a}$) with LightNN-1, we find that FLightNNs can achieve higher accuracy with the same or even lower storage as LightNN-1. This is because initially FLightNNs quantize all the filters with two shifts (since $\mathbf{t}$ is initialized as $\mathbf{0}$), and gradually add constraints to the filters. This gradual quantization may be better than training a network with only one shift from scratch, as LightNN-1 does. The benefit of gradual quantization has also been observed by prior work~\cite{dong2017learning} which shows that gradually imposing quantization constraints can achieve better accuracy than directly quantizing with a strict constraint. 

Table~\ref{FPGA_resource_utilization} shows the FPGA resources utilization for networks 7 and 8. Since full-precision and fixed-point DNNs require DSP for both multiplication and addition, while LightNNs and FLightNNs only need DSP for addition, full-precision and fixed-point DNNs have larger DSP resource utilization. Compared to full-precision DNNs which use 32-bit floating point operations, fixed-point DNNs only use 4-bit weights and 8-bit activations, and therefore consume fewer DSP units. LightNNs and FLightNNs use LUT to implement the multipliers, and have a higher utilization of LUT than full-precision and fixed-point DNNs. However, the performance of (F)LightNNs is not bounded by LUT resources since the maximum usage of LUT by LightNN-2 is only 42\% and 17\% for networks 7 and 8, respectively. Instead, the memory resource (BRAM) bounds the performance for (F)LightNNs, while for full-precision and fixed-point DNNs, the performance is bounded by both BRAM and DSP. 

\begin{table}[t]
\centering
\caption{Accuracy and FPGA throughput for CIFAR-10. In the ``Model" column, ``Full", ``L-2", ``L-1", ``FP", ``FL" indicate full-precision DNN, LightNN-2, LightNN-1, Fixed-point DNN, and FLightNN, respectively. The subscript ``$x$W$y$A" indicates $x$ bits for weights and $y$ bits for activations. The FLightNN results are shown in bold face. We use subscript $a$ and $b$ to denote the two trained FLightNNs for each network. These notations also apply for Table~\ref{table:results_svhn}, \ref{table:results_cifar100} and \ref{table:results_imagenet}.}
\vspace{-12pt}
\label{table:results_cifar10}
\begin{tabular}{C{0.25cm}|C{1.2cm}|C{1.1cm}|C{0.9cm}|C{1.5cm}|C{1cm}}
    \hline
    ID & Model & Accuracy (\%) & Storage (MB) & Throughput (images/s) & Speedup \\
    \hline
    \multirow{6}{*}{1} & Full & 86.36 & 0.31 & 3.2e2 & 1$\times$ \\
    \hhline{~-----}
    & L-2$_{8W8A}$ & 86.17 & 0.08 & 2.2e3 & 7.0$\times$ \\
    \hhline{~-----}
    & L-1$_{4W8A}$ & 84.82 & 0.04 & 4.5e3 & 14.4$\times$ \\
    \hhline{~-----}
    & FP$_{4W8A}$ & 85.09 & 0.04 & 3.3e3 & 10.5$\times$ \\
    \hhline{~-----}
    & \textbf{FL$_{1a}$} & \textbf{85.70} & \textbf{0.04} & \textbf{4.8e3} & \textbf{15.0$\times$} \\
    \hhline{~-----}
    & \textbf{FL$_{1b}$} & \textbf{85.91} & \textbf{0.06} &  \textbf{4.0e3} & \textbf{12.6$\times$} \\
    \hline
    \multirow{6}{*}{2} & Full & 91.70 & 2.8 & 1.4e2 & 1$\times$ \\
    \hhline{~-----}
    & L-2$_{8W8A}$ & 91.64 & 0.7 & 1.6e3 & 11.5$\times$ \\
    \hhline{~-----}
    & L-1$_{4W8A}$ & 91.15 & 0.4 & 2.7e3 & 19.0$\times$ \\
    \hhline{~-----}
    & FP$_{4W8A}$ & 91.17 & 0.4 & 1.5e3 & 10.7$\times$ \\
    \hhline{~-----}
    & \textbf{FL$_{2a}$} & \textbf{91.36} & \textbf{0.4} & \textbf{2.8e3} & \textbf{18.9$\times$} \\
    \hhline{~-----}
    & \textbf{FL$_{2b}$} & \textbf{91.48} & \textbf{0.7} & \textbf{1.9e3} & \textbf{13.0$\times$} \\
    \hline
    \multirow{6}{*}{3} & Full & 92.85 & 18.5 & 1.3 & 1$\times$ \\
    \hhline{~-----}
    & L-2$_{8W8A}$ & 92.72 & 4.6 & 10.2 & 7.8$\times$ \\
    \hhline{~-----}
    & L-1$_{4W8A}$ & 91.93 & 2.3 & 39.2 & 30.2$\times$ \\
    \hhline{~-----}
    & FP$_{4W8A}$ & 92.23 & 2.3 & 19.8 & 15.2$\times$ \\
    \hhline{~-----}
    & \textbf{FL$_{3a}$} & \textbf{92.59} & \textbf{2.3} & \textbf{39.2} & \textbf{30.2$\times$} \\
    \hhline{~-----}
    & \textbf{FL$_{3b}$} & \textbf{92.62} & \textbf{3.3} & \textbf{27.2} & \textbf{21.0$\times$} \\
    \hline
\end{tabular}
\end{table}

\begin{table}[t]
\centering
\caption{Accuracy and FPGA throughput for SVHN.}
\label{table:results_svhn}
\begin{tabular}{C{0.25cm}|C{1.2cm}|C{1.1cm}|C{0.9cm}|C{1.5cm}|C{1cm}}
    \hline
    ID & Model & Accuracy (\%) & Storage (MB) & Throughput (images/s) & Speedup \\
    \hline
    \multirow{6}{*}{4} & Full & 94.96 & 0.12 & 2.2e3 & 1$\times$ \\
    \hhline{~-----}
    & L-2$_{8W8A}$ & 94.90 & 0.03 & 4.5e3 & 2.09$\times$ \\
    \hhline{~-----}
    & L-1$_{4W8A}$ & 94.16 & 0.02 & 8.3e3 & 3.63$\times$ \\
    \hhline{~-----}
    & FP$_{4W8A}$ & 93.70 & 0.02 & 3.7e3 & 1.70$\times$ \\
    \hhline{~-----}
    & \textbf{FL$_{4a}$} & \textbf{94.67} & \textbf{0.02} & \textbf{6.7e3} & \textbf{3.11$\times$} \\
    \hhline{~-----}
    & \textbf{FL$_{4b}$} & \textbf{94.88} & \textbf{0.03} & \textbf{5.3e3} & \textbf{2.37$\times$} \\
    \hline
    \multirow{6}{*}{5} & Full & 96.44 & 0.4 & 1.1e3 & 1$\times$ \\
    \hhline{~-----}
    & L-2$_{8W8A}$ & 96.38 & 0.1 & 2.1e3 & 2.00$\times$ \\
    \hhline{~-----}
    & L-1$_{4W8A}$ & 95.93 & 0.05 & 3.7e3 & 3.53$\times$ \\
    \hhline{~-----}
    & FP$_{4W8A}$ & 96.02 & 0.05 & 1.8e3 & 1.71$\times$ \\
    \hhline{~-----}
    & \textbf{FL$_{5a}$} & \textbf{96.21} & \textbf{0.06} & \textbf{3.2e3} & \textbf{3.06$\times$} \\
    \hhline{~-----}
    & \textbf{FL$_{5b}$} & \textbf{96.24} & \textbf{0.08} & \textbf{3.0e3} & \textbf{2.84$\times$} \\
    \hline
\end{tabular}
\end{table}

\begin{table}[t]
\centering
\caption{Accuracy and FPGA throughput for CIFAR-100.}
\label{table:results_cifar100}
\begin{tabular}{C{0.25cm}|C{1.2cm}|C{1.1cm}|C{0.9cm}|C{1.5cm}|C{1cm}}
    \hline
    ID & Model & Accuracy (\%) & Storage (MB) & Throughput (images/s) & Speedup \\
    \hline
    \multirow{6}{*}{6} & Full & 69.16 & 2.8 & 2.5e2 & 1$\times$ \\
    \hhline{~-----}
    & L-2$_{8W8A}$ & 68.84 & 0.7 & 1.6e3 & 6.4$\times$ \\
    \hhline{~-----}
    & L-1$_{4W8A}$ & 67.32 & 0.4 & 2.7e3 & 10.6$\times$ \\
    \hhline{~-----}
    & FP$_{4W8A}$ & 67.67 & 0.4 & 1.5e3 & 5.98$\times$ \\
    \hhline{~-----}
    & \textbf{FL$_{6a}$} & \textbf{68.59} & \textbf{0.4} & \textbf{2.7e3} & \textbf{10.6$\times$} \\
    \hhline{~-----}
    & \textbf{FL$_{6b}$} & \textbf{68.76} & \textbf{0.6} & \textbf{1.8e3} & \textbf{6.88$\times$} \\
    \hline
    
    \multirow{6}{*}{7} & Full & 71.22 & 11.2 & 7.4e1 & 1$\times$ \\
    \hhline{~-----}
    & L-2$_{8W8A}$ & 70.96 & 2.8 & 6.0e2 &8.11$\times$ \\
    \hhline{~-----}
    & L-1$_{4W8A}$ & 69.71 & 1.4 & 1.1e3 & 15.2$\times$ \\
    \hhline{~-----}
    & FP$_{4W8A}$ & 69.34 & 1.4 & 6.9e2 & 9.26$\times$ \\
    \hhline{~-----}
    & \textbf{FL$_{7a}$} & \textbf{70.85} & \textbf{1.4} & \textbf{1.1e3} & \textbf{15.2$\times$} \\
    \hhline{~-----}
    & \textbf{FL$_{7b}$} & \textbf{70.87} & \textbf{2.4} & \textbf{7.4e2} & \textbf{9.98$\times$} \\
    \hline
\end{tabular}
\end{table}
     
\begin{table}[t]
\centering
\caption{Top-5 Accuracy and FPGA throughput for ImageNet.}
\label{table:results_imagenet}
\begin{tabular}{C{0.25cm}|C{1.2cm}|C{1.1cm}|C{0.9cm}|C{1.5cm}|C{1cm}}
    \hline
    ID & Model & Accuracy (\%) & Storage (MB) & Throughput (images/s) & Speedup \\
    \hline
    \multirow{4}{*}{8} & L-2$_{8W8A}$ & 75.04 & 1.8 & 2.7e2 & 1$\times$ \\
    \hhline{~-----}
    & L-1$_{4W8A}$ & 72.94 & 0.9 & 5.2e2 & 1.95$\times$ \\
    \hhline{~-----}
    & \textbf{FL$_{8a}$} & \textbf{74.80} & \textbf{1.5} & \textbf{3.1e2} & \textbf{1.16$\times$} \\
    \hhline{~-----}
    & \textbf{FL$_{8b}$} & \textbf{75.00} & \textbf{1.7} & \textbf{2.8e2} & \textbf{1.06$\times$} \\
    \hline
\end{tabular}
\end{table}

\begin{table}[t]
\caption{FPGA resource utilization for different quantized DNN models.}
\label{FPGA_resource_utilization}
\begin{center} 
 \begin{tabular}{ C{0.25cm} | C{1.2cm} | C{0.85cm} | C{0.85cm} | C{0.85cm} | C{0.85cm} | C{1.1cm} }
 \hline
  \multirow{2}{*}{ID} & \multirow{2}{*}{Model} & \multicolumn{4}{c|}{Maximum resource utilization} & \multirow{2}{*}{Speedup}\\ 
 \hhline{~~----~}
  & & BRAM & DSP & FF & LUT &   \\
 \hline\hline
  \multirow{6}{*}{7} & Full & 896 & 642 & 69,344 & 128,339 & 1$\times$ \\
  \hhline{~------}
  & L-2$_{8W8A}$ &  1,024 & 4  &  66,491 &  90,949& 8.11$\times$ \\
  \hhline{~------}
  & L-1$_{4W8A}$ &  1,024 & 4  &  66,491 &  90,949& 15.2$\times$\\
  \hhline{~------}
  & FP$_{4W8A}$ &  1,024 & 514  &  7,110 &  90,949& 9.26$\times$ \\
  \hhline{~------}
  & \textbf{FL$_{7a}$} &   1,024 & 4  &  84,192 &  90,949& 9.98$\times$ \\
  \hhline{~------}
  &\textbf{FL$_{7b}$} &  1,024 & 4  &  63,648 &  80,940& 15.2$\times$ \\
 \hline
 
  \multirow{4}{*}{8} & L-2$_{8W8A}$ & 832 & 16 & 3,156 & 38,022& 1$\times$ \\
  \hhline{~------}
  & L-1$_{8W8A}$  & 800 & 16 & 3,084 & 36,906& 1.95$\times$ \\
  \hhline{~------}
   & \textbf{FL$_{8a}$} &  800 & 16 & 5,070 & 36,098& 1.16$\times$ \\
  \hhline{~------}
  &\textbf{FL$_{8b}$} &800 & 16 & 6,272 & 37,406& 1.06$\times$ \\
 \hline
 \multicolumn{2}{c|}{Available} &  1,090& 900& 437,200 & 218,600& \\ 
  \hline
\end{tabular} 
\end{center}
\end{table}

\subsection{Accuracy-energy trade-off on ASIC}
For all quantized DNNs, we designed pipelined implementations with one stage per neuron, where the computation unit is reused for each neuron. A 65nm commercial standard library is adopted. The Synopsys Design Compiler~\cite{manual2010synopsys} is used to generate the gate-level netlist of the computation units.  The power consumption of all computation operations within one layer is calculated using Synopsys Primetime. We keep all the DNN architectures implemented in an unoptimized fashion because our main objective is to compare how different quantized DNNs impact computational energy. 

\begin{figure}[t]
  \vspace{-12pt}
  \centering
  \includegraphics[width=0.23\textwidth, height=2.7cm]{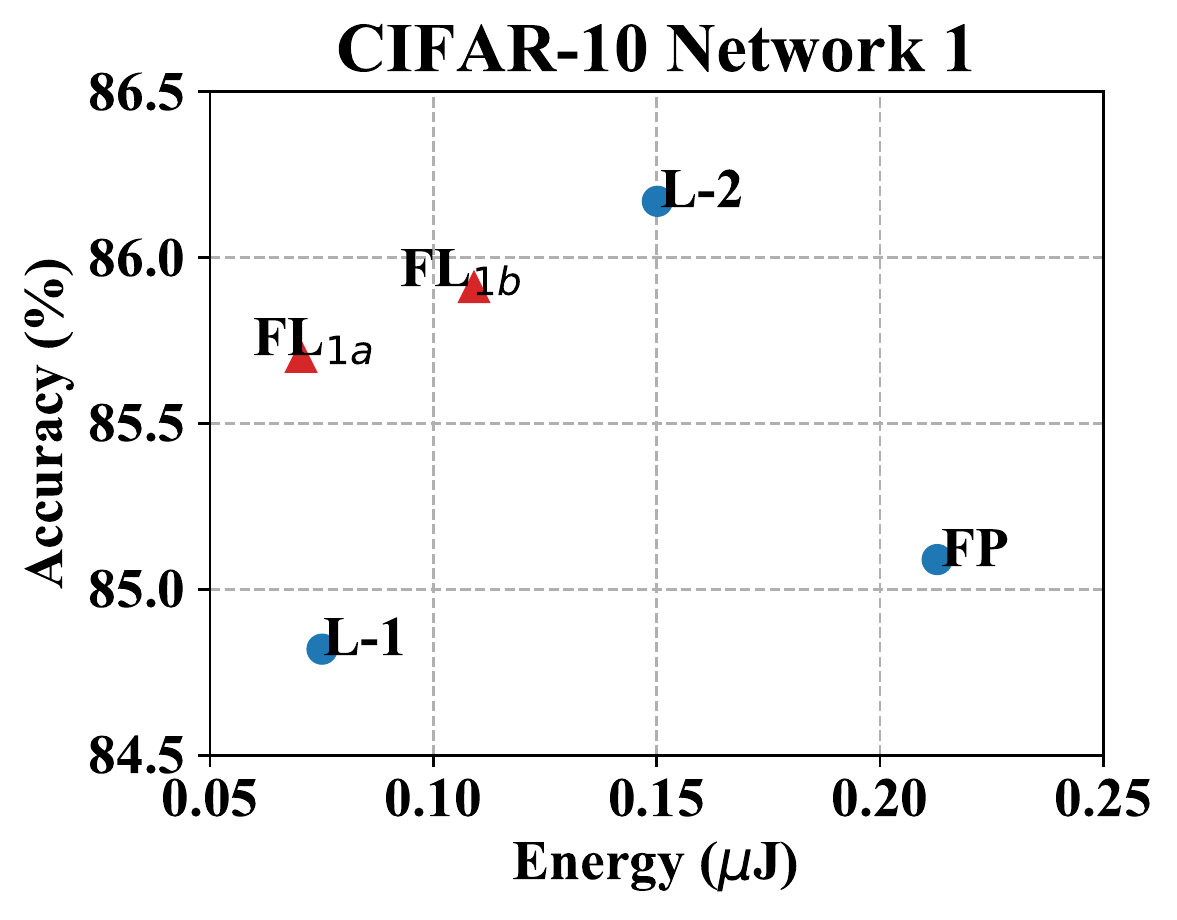}
  \includegraphics[width=0.23\textwidth, height=2.7cm]{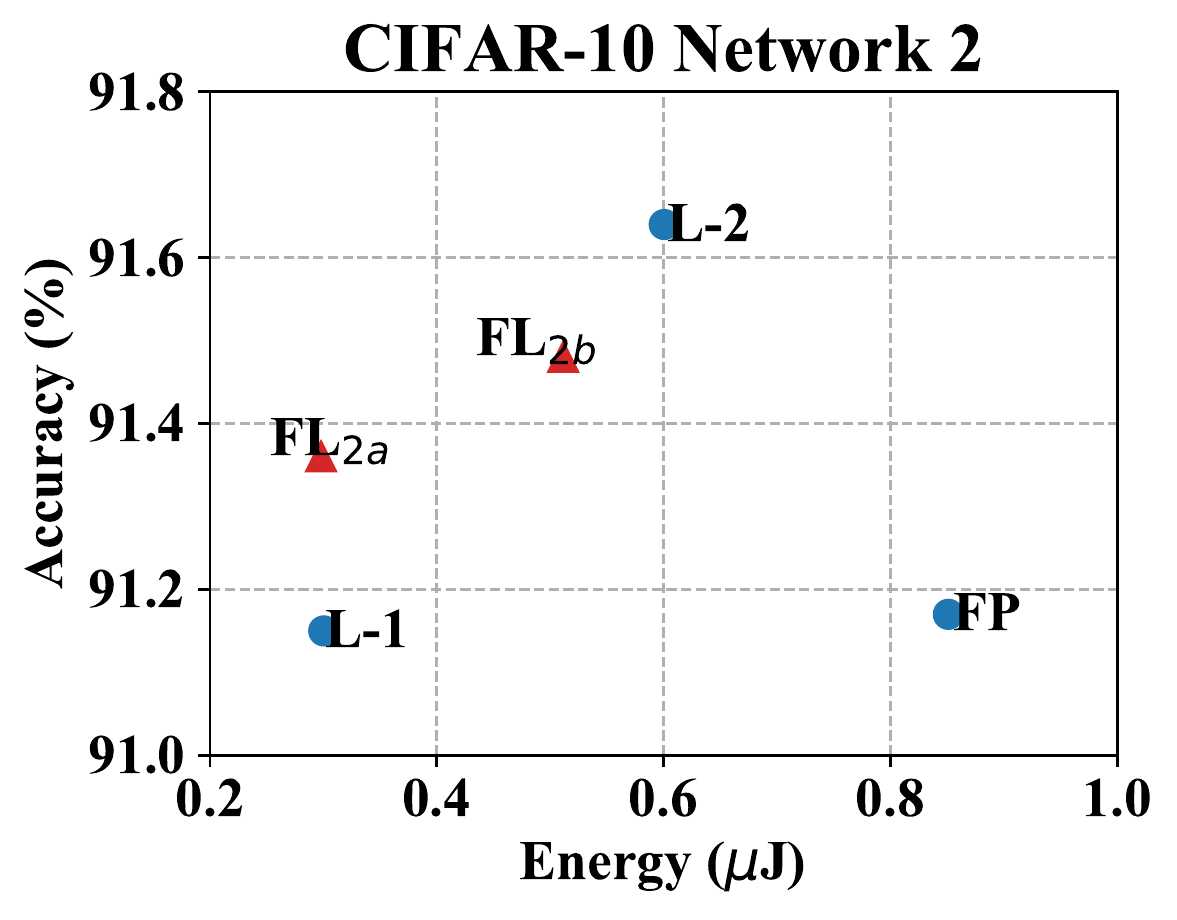}\\
  \includegraphics[width=0.23\textwidth, height=2.7cm]{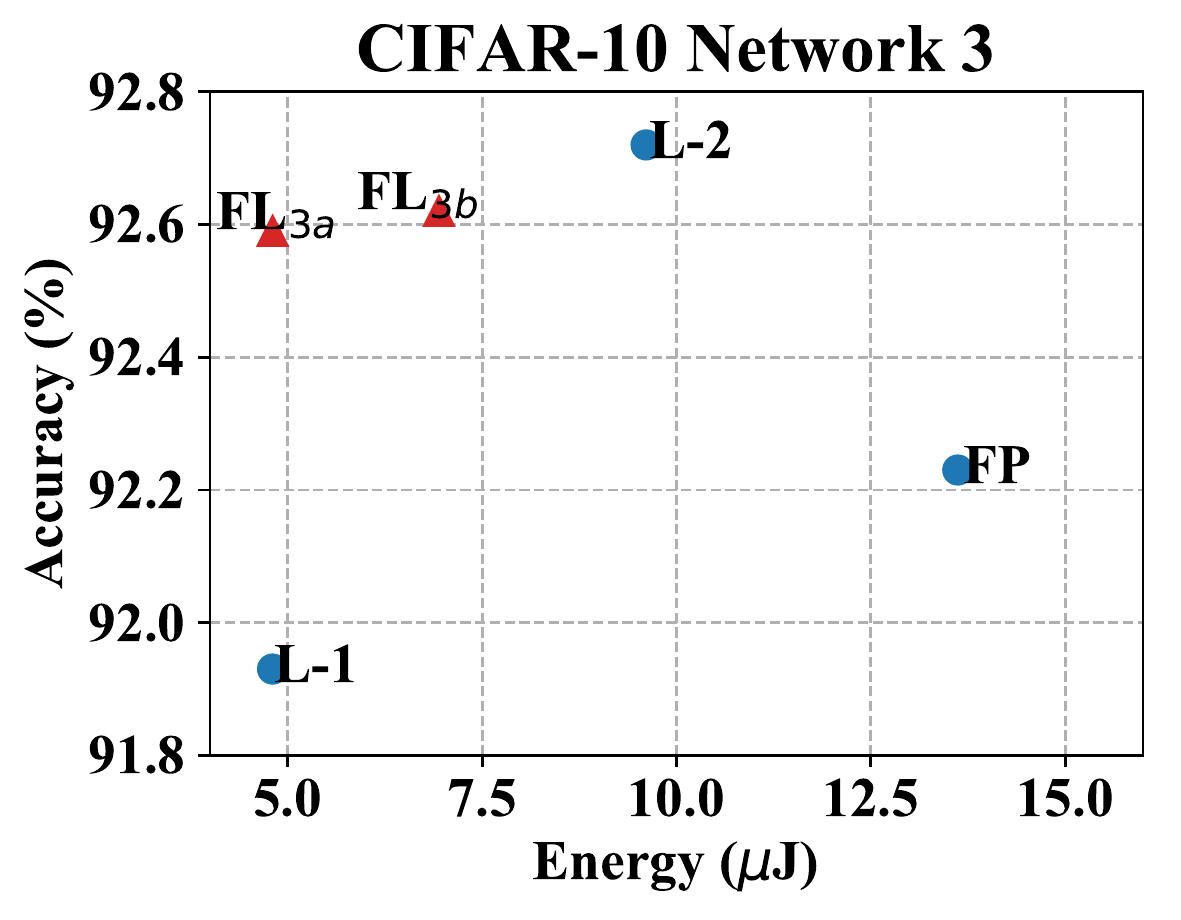}
  \includegraphics[width=0.23\textwidth, height=2.7cm]{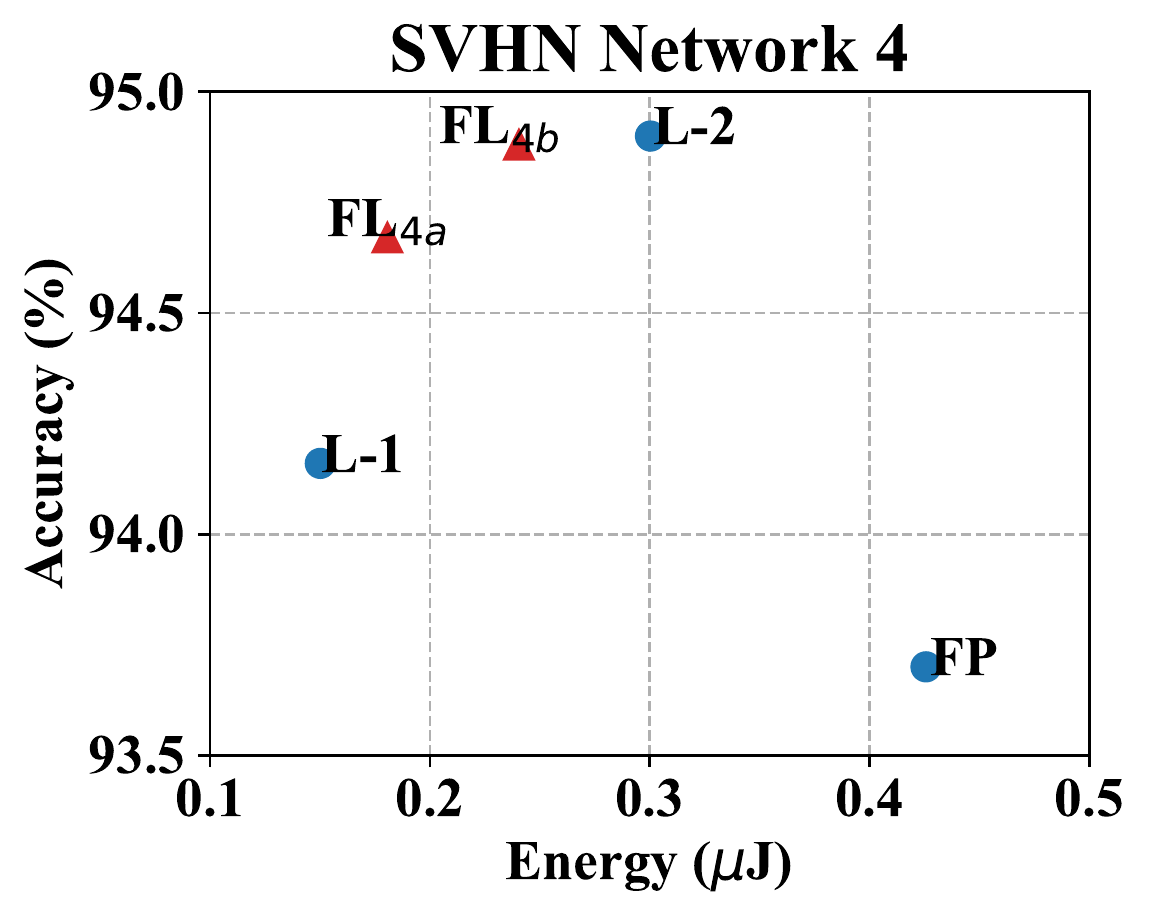}\\
  \includegraphics[width=0.23\textwidth, height=2.7cm]{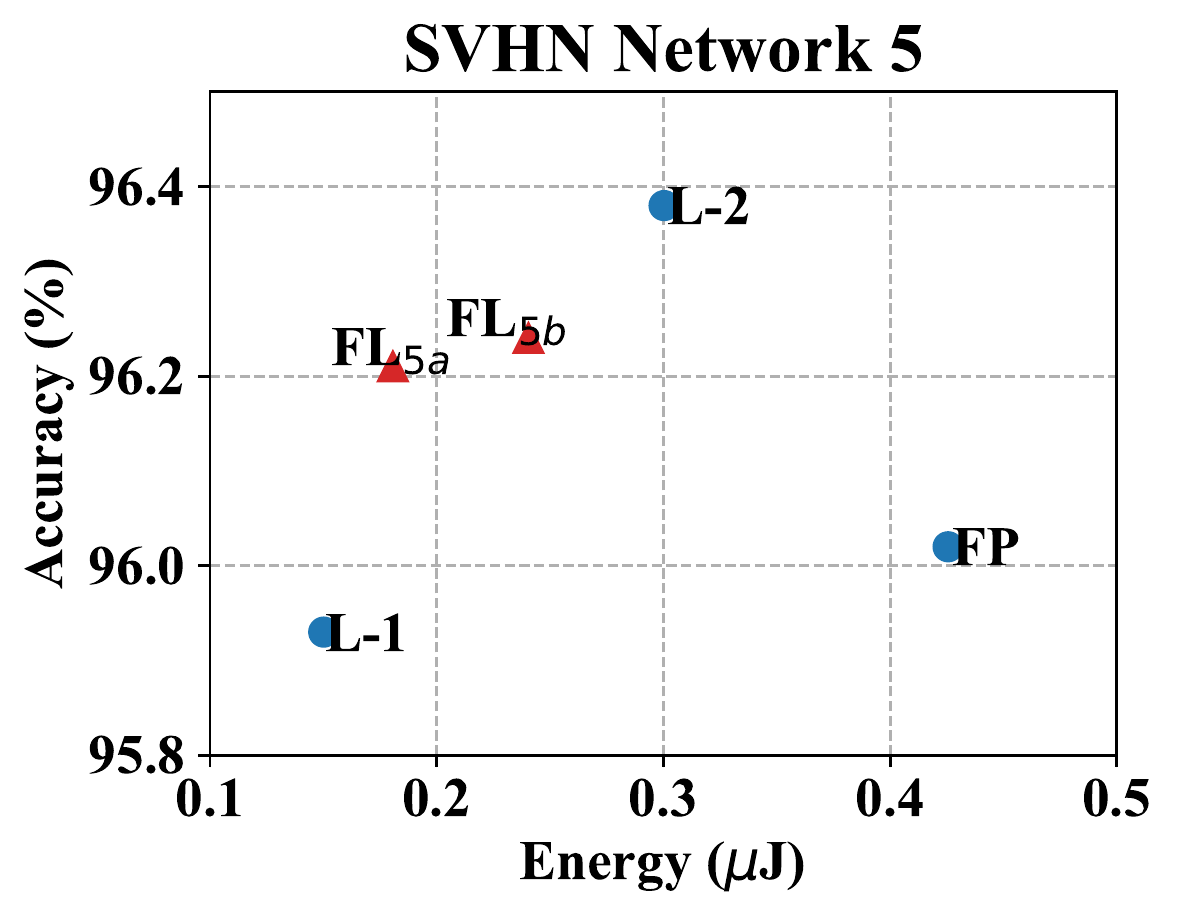}
  \includegraphics[width=0.23\textwidth, height=2.7cm]{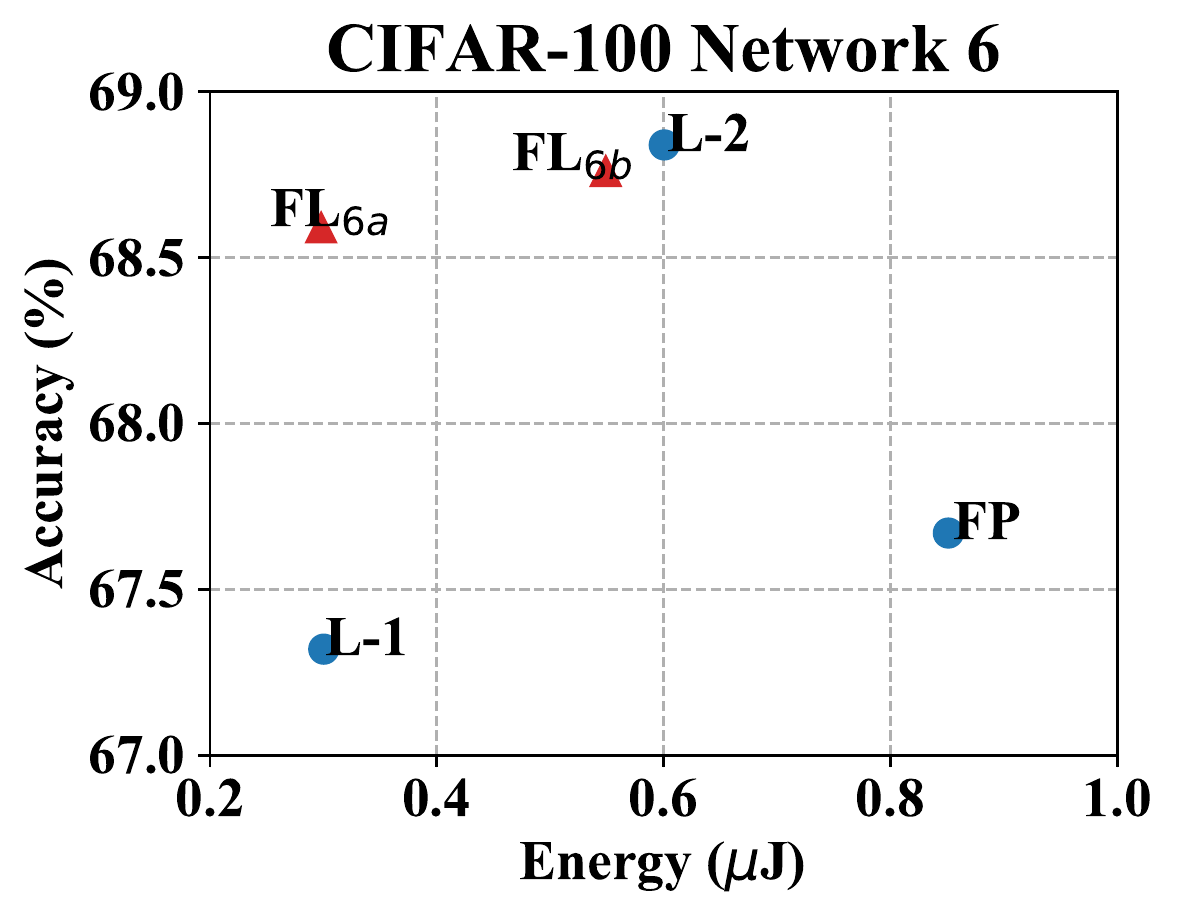}\\
  \includegraphics[width=0.23\textwidth, height=2.7cm]{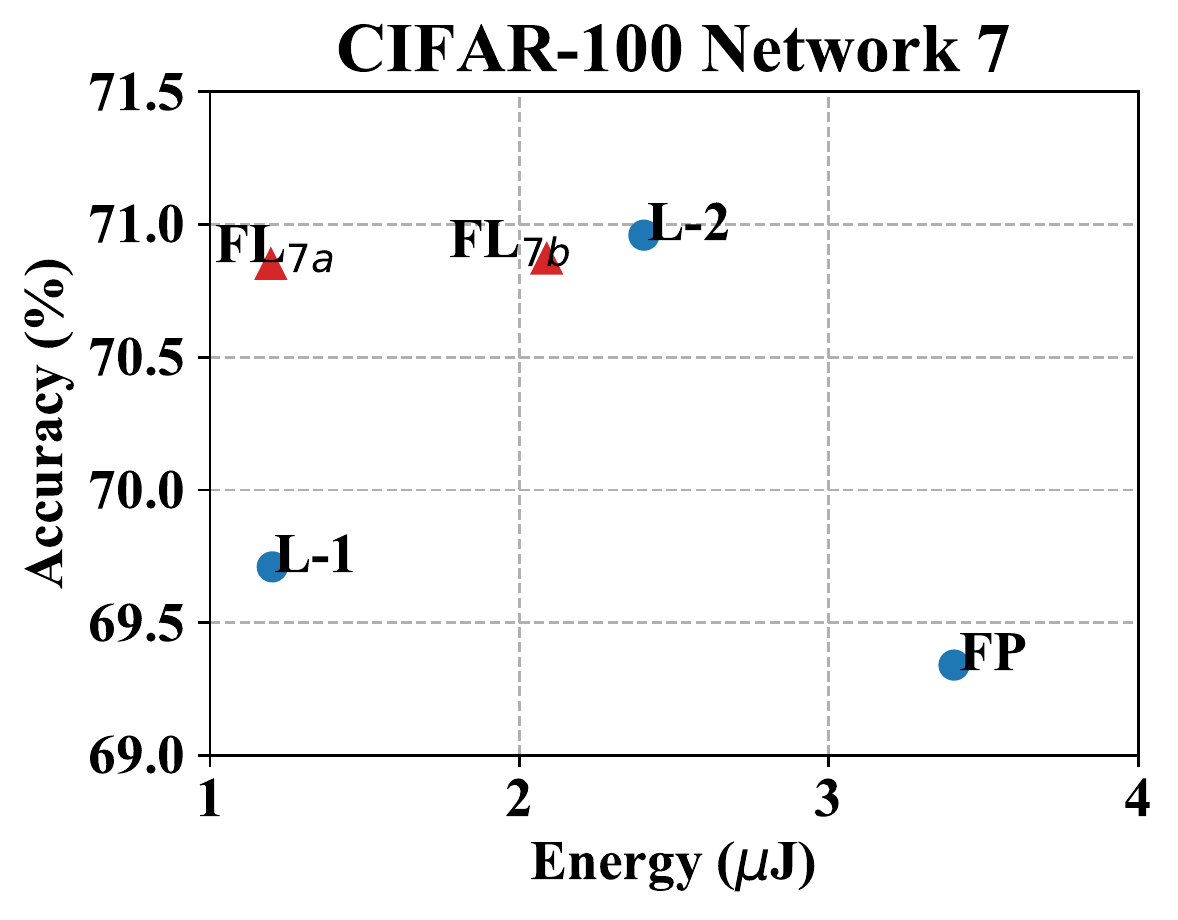}
  \includegraphics[width=0.23\textwidth, height=2.7cm]{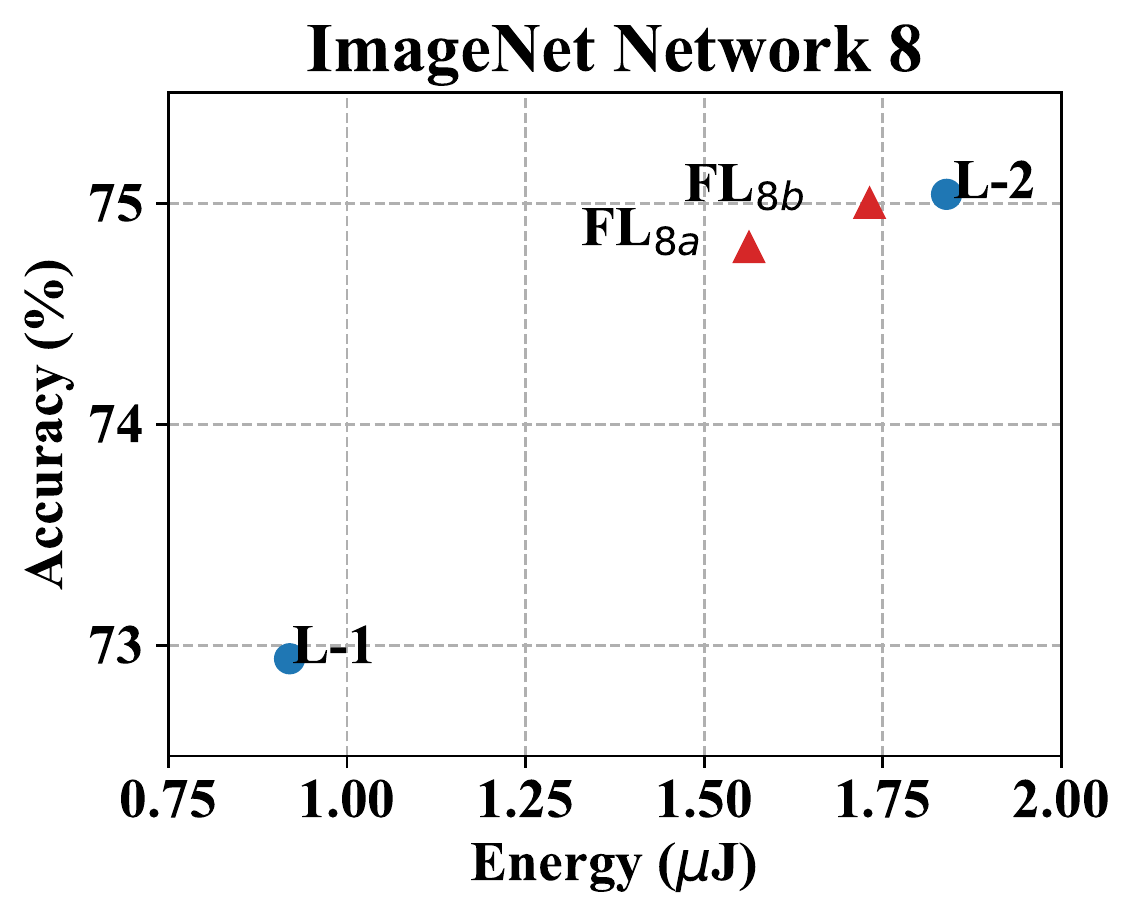}\\
  \caption{Accuracy and computational energy consumption in ASIC for different quantized models on CIFAR-10, SVHN, CIFAR-100 and ImageNet datasets. FLightNNs are marked as red triangles, while the other models are shown as blue dots.}\label{fig:pareto-asic}
\end{figure}

The accuracy and computational energy trade-offs for the quantized DNN models are shown in Fig.~\ref{fig:pareto-asic}. The energy shown in Fig.~\ref{fig:pareto-asic} only includes the computational energy consumption for the largest layer of each network. We can clearly observe that FLightNNs provide a more continuous Pareto front for LightNN-2 and LightNN-1, regardless of the network type (\textit{i.e.}, VGG or ResNet), size and the datasets.  Similar to the observation in Sec.~\ref{sec:fpga}, in some networks FLightNNs can achieve higher accuracy than LightNN-1 with lower computational energy cost. 

\section{Discussion}

\begin{figure}[t]
  \centering
  \includegraphics[width=0.22\textwidth,height=2.3cm]{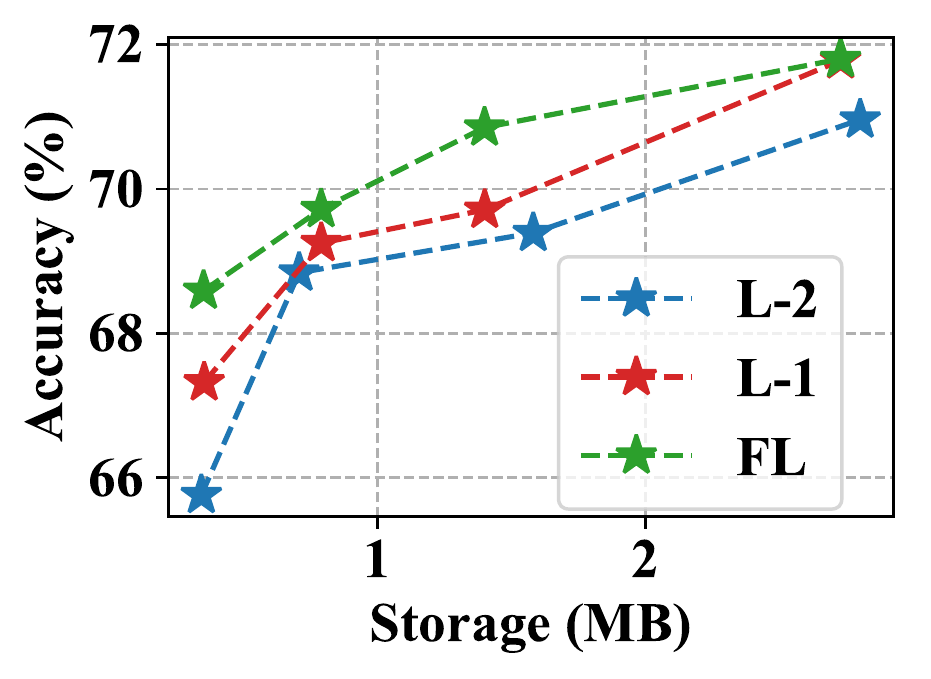}
  \caption{Accuracy-storage front for LightNN-2 LightNN-1 and FLightNN. The Pareto front of FLightNN is the upper bound of LightNNs.}\label{fig:L2-L1-FL-frontier}
\end{figure}

Since FLightNNs customize the $\mathbf{k}_i$ for each filter, LightNN-1 and LightNN-2 can be considered as two special cases for FLightNNs. Therefore, the Pareto front created by the searched FLightNN solutions should be the upper bound for the front of LightNN-1 and LightNN-2 with varied parameter numbers. We test this hypothesis on CIFAR-100 dataset using networks with varied number of convolutional filters. As shown in Fig.~\ref{fig:L2-L1-FL-frontier}, the accuracy-storage Pareto-front created by FLightNNs is consistently higher than the LightNNs. This indicates that instead of only filling in the Pareto front of LightNNs, FLightNNs can \textit{push forward} the Pareto front, due to their larger design space. The proposed differentiable training algorithm optimizes both $\mathbf{k}_i$ and weight values in an end-to-end fashion, and therefore significantly reduces searching effort compared to exhaustive or heuristic methods with multiple rounds of training. Future work will further improve training efficiency by using optimized training loss~\cite{ding2019regularizing} or proper labels~\cite{chen2018understanding}. 

\section{Conclusion}

In this paper, we propose FLightNNs which customize the number of shift operations for each filter of LightNNs. Equipped with the proposed differentiable training algorithm, FLightNNs can achieve a flexible trade-off between accuracy and speed/energy. Our experimental results on FPGA and ASIC simulations show that FLightNNs can provide a more continuous Pareto front for LightNN models and consistently outperform fixed-point DNNs \textit{w.r.t.} both accuracy and speed/energy. Moreover, due to the gradual quantization nature of the differentiable training, FLightNNs can achieve higher accuracy than LightNN-1 without sacrificing speed and energy efficiency, and thus, push forward the Pareto-optimal front. These promising results suggest the potentials for FLightNNs to achieve fast and accurate inference on learning-based customized hardware. 

\begin{acks}
This research was supported in part by NSF CCF Grant No. 1815899.
\end{acks}

{\tiny
\bibliographystyle{ACM-Reference-Format}

\bibliography{dac2019}
}

\end{document}